\documentclass[11pt]{cambrian}
\usepackage[utf8]{inputenc}
\DeclareUnicodeCharacter{207A}{\textsuperscript{+}}
\usepackage{graphicx}
\usepackage{multirow}    
\usepackage{booktabs}    
\usepackage{float}       
\usepackage{amsmath}      
\usepackage{subcaption}   
\usepackage{tcolorbox}    
\usepackage{xcolor}
\tcbset{floatplacement=t}

\usepackage{cleveref}

\usepackage{xspace}

\newcommand{\eg}[0]{\emph{e.g., }}

\usepackage{subcaption}   

\makeatletter\renewcommand\paragraph{\@startsection{paragraph}{4}{\z@}
{.2em \@plus1ex \@minus.2ex}{-.5em}{\normalfont\normalfont\bfseries}}
\makeatother

\newcolumntype{x}[1]{>{\centering\arraybackslash}p{#1pt}}
\newcolumntype{y}[1]{>{\raggedright\arraybackslash}p{#1pt}}
\newcolumntype{z}[1]{>{\raggedleft\arraybackslash}p{#1pt}}

\newcommand{\model}{\mbox{\texttt{\textbf{PyVision}}\xspace}}

\newcommand{\crop}{\mbox{\textsc{{Crop}}\xspace}}
\newcommand{\rotatetool}{\mbox{\textsc{{Rotate}}\xspace}}
\newcommand{\enhance}{\mbox{\textsc{{Enhance Contrast}}\xspace}}
\newcommand{\seg}{\mbox{\textsc{{Segmentation}}\xspace}}
\newcommand{\detec}{\mbox{\textsc{{Detection}}\xspace}}
\newcommand{\ocr}{\mbox{\textsc{{OCR}}\xspace}}
\newcommand{\rendermarks}{\mbox{\textsc{{Render Marks}}\xspace}}
\newcommand{\renderlines}{\mbox{\textsc{{Render Auxiliary Lines}}\xspace}}
\newcommand{\hist}{\mbox{\textsc{{Visulize Image Histogram}}\xspace}}
\newcommand{\numericalana}{\mbox{\textsc{{Numerical Analysis}}\xspace}}

\crefname{table}{Tab.}{Tabs.}
\Crefname{table}{Tab.}{Tabs.}

\crefname{figure}{Fig.}{Figs.}
\Crefname{figure}{Fig.}{Figs.}

\crefname{section}{Sec.}{Secs.}
\Crefname{section}{Sec.}{Secs.}

\crefname{equation}{Eq.}{Eqs.}
\Crefname{equation}{Eq.}{Eqs.}

\newcommand{\toolbasic}[1]{\textbf{\textcolor[RGB]{81,151,199}{#1}}}
\newcommand{\tooladvanced}[1]{\textbf{\textcolor[RGB]{189,85,81}{#1}}}
\newcommand{\toolprompts}[1]{\textbf{\textcolor[RGB]{100,169,86}{#1}}}
\newcommand{\toolnumerical}[1]{\textbf{\textcolor[RGB]{219,156,73}{#1}}}

\newcommand{\scorediff}[2]{\makecell[c]{#1\\[-2pt] \textcolor{green!70!black}{#2}}}
\usepackage[utf8]{inputenc}

\usepackage[numbers,sort&compress]{natbib}
\usepackage{colortbl}

\usepackage[utf8]{inputenc} 
\usepackage[T1]{fontenc}    
\usepackage{hyperref}
\usepackage{url}            
\usepackage{booktabs}       
\usepackage{amsfonts}       
\usepackage{nicefrac}       
\usepackage{microtype}      
\usepackage{xcolor}
\definecolor{bluelink}{RGB}{0,113,188}
\definecolor{greenlink}{RGB}{0,188,113}
\hypersetup{
    colorlinks=true,%
    citecolor=green!93!black,%
    filecolor=redlink,%
    linkcolor=red!93!black,%
    urlcolor=bluelink
}
\usepackage{tabularx}
\usepackage{tcolorbox}
\usepackage{amsmath}
\usepackage{multirow}
\usepackage{array}
\usepackage{caption}
\usepackage{wrapfig}
\usepackage{enumitem}
\usepackage{tikz}
\usepackage{lipsum}
\usepackage{subcaption}
\usepackage{multirow} 
\usepackage{adjustbox}

\usepackage{amssymb}
\usepackage{unicode}  

\usepackage{pgf}
\usepackage{colortbl}

\usepackage{tipa}

\usepackage{rotating}
\usepackage[abs]{overpic}
\usepackage{makecell}
\usepackage{longtable}

\usepackage{tocloft}  

\usepackage[english]{babel}
\usepackage{csquotes}
\usepackage{listings}
\definecolor{codekeyword}{rgb}{0.0, 0.0, 0.5}   
\definecolor{codecomment}{rgb}{0.0, 0.5, 0.0}   
\definecolor{codestring}{rgb}{0.56, 0.0, 1.0}   
\definecolor{backcolour}{rgb}{0.98,0.98,0.97}

\lstdefinestyle{pythonstyle}{
    language=Python,                          
    basicstyle=\ttfamily\small,               
    keywordstyle=\color{codekeyword}\bfseries,
    commentstyle=\color{codecomment}\itshape, 
    stringstyle=\color{codestring},           
    showstringspaces=false,                   
    breaklines=true,                          
    tabsize=4,                                
    numbers=none,                             
    frame=single, 
    backgroundcolor=\color{backcolour},
    captionpos=b,                             
    morekeywords={self, __init__, __name__, __main__}, 
}

\usepackage{fontawesome}
\usepackage{xspace}
\newcommand{\huggingface}{\raisebox{-1.5pt}{\includegraphics[height=1.05em]{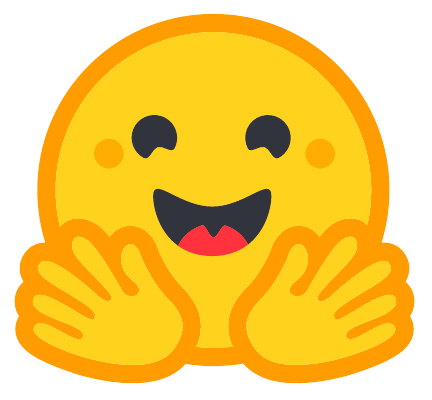}}\xspace}
\newcommand{\github}{\raisebox{-1.5pt}{\includegraphics[height=1.05em]{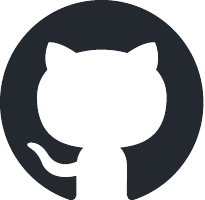}}\xspace}
\newcommand{\worldwideweb}{\raisebox{-1.5pt}{\includegraphics[height=1.05em]{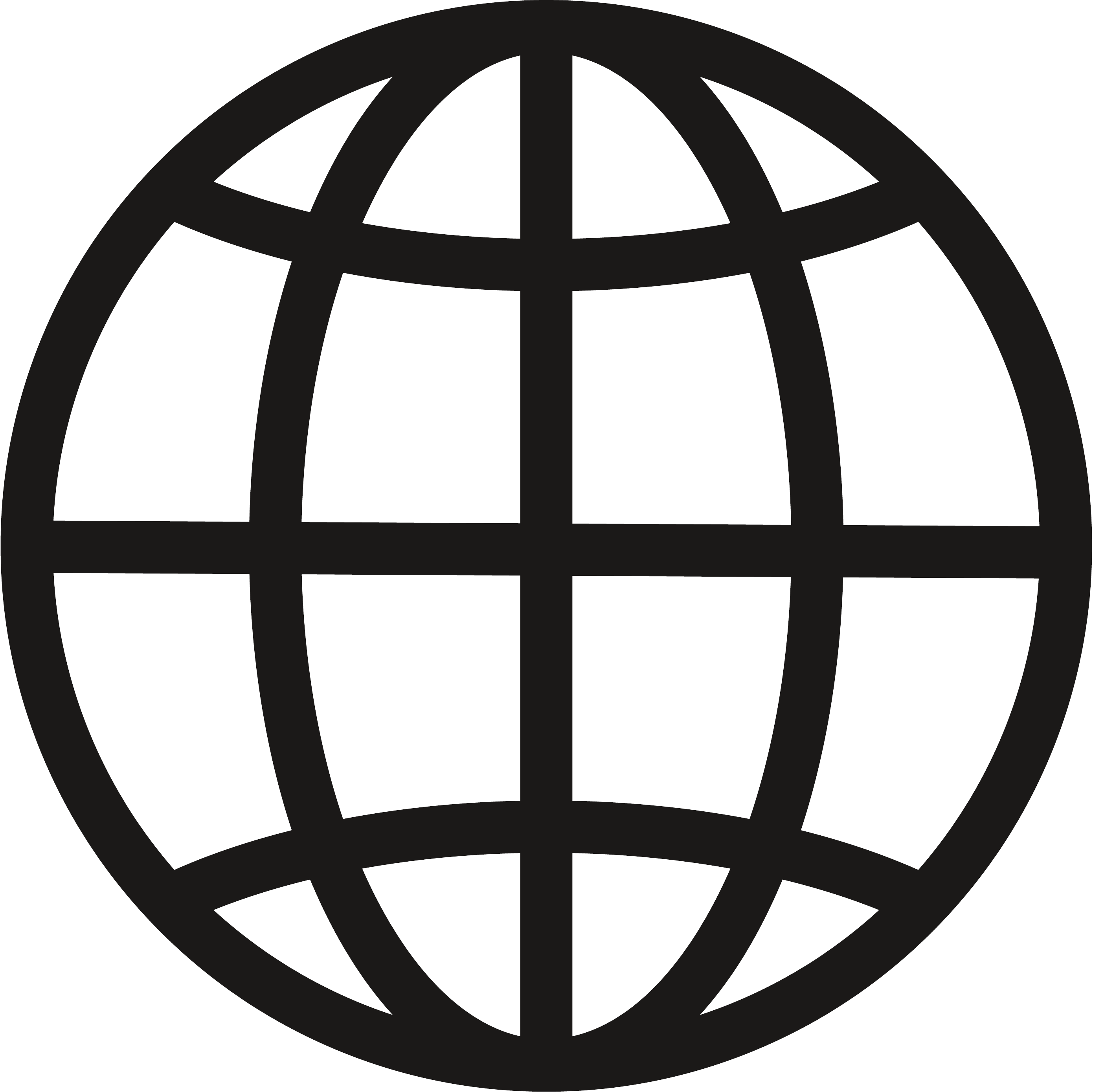}}\xspace}

\title{\center \model: Agentic Vision with Dynamic Tooling}

\renewcommand{\thefootnote}{\fnsymbol{footnote}}

\author{%
Shitian~Zhao\textsuperscript{1,\textsection,}\footnote{Joint First Author; \textsuperscript{\textsection}Project Lead; 
\textsuperscript{\textdagger}Corresponding Author}, 
Haoquan~Zhang\textsuperscript{1,3,*}, 
Shaoheng~Lin\textsuperscript{1,*}, 
Ming~Li\textsuperscript{1,*}, 
Qilong~Wu\textsuperscript{4,*}, 
Kaipeng~Zhang\textsuperscript{1,5,\textdagger},
Chen~Wei\textsuperscript{2,\textdagger}\\
\textsuperscript{1}Shanghai AI Lab,\quad 
\textsuperscript{2}Rice University,\quad 
\textsuperscript{3}CUHK,\quad 
\textsuperscript{4}NUS,\quad 
\textsuperscript{5}SII\\

\vspace{5pt}

{\worldwideweb \href{https://agent-x.space/pyvision/}{{\text{Project Page}}}} \quad \quad {\github \href{https://github.com/agents-x-project/PyVision}{{\text{Inference Code}}}}
\quad \quad
{\huggingface \href{https://huggingface.co/spaces/Agents-X/PyVision}{{\text{PyVision Demo}}}}
}

\begin{document}

\maketitle

\begin{figure}[h]
    \vspace{-1.2em}
    \centering
    \includegraphics[width=\textwidth]{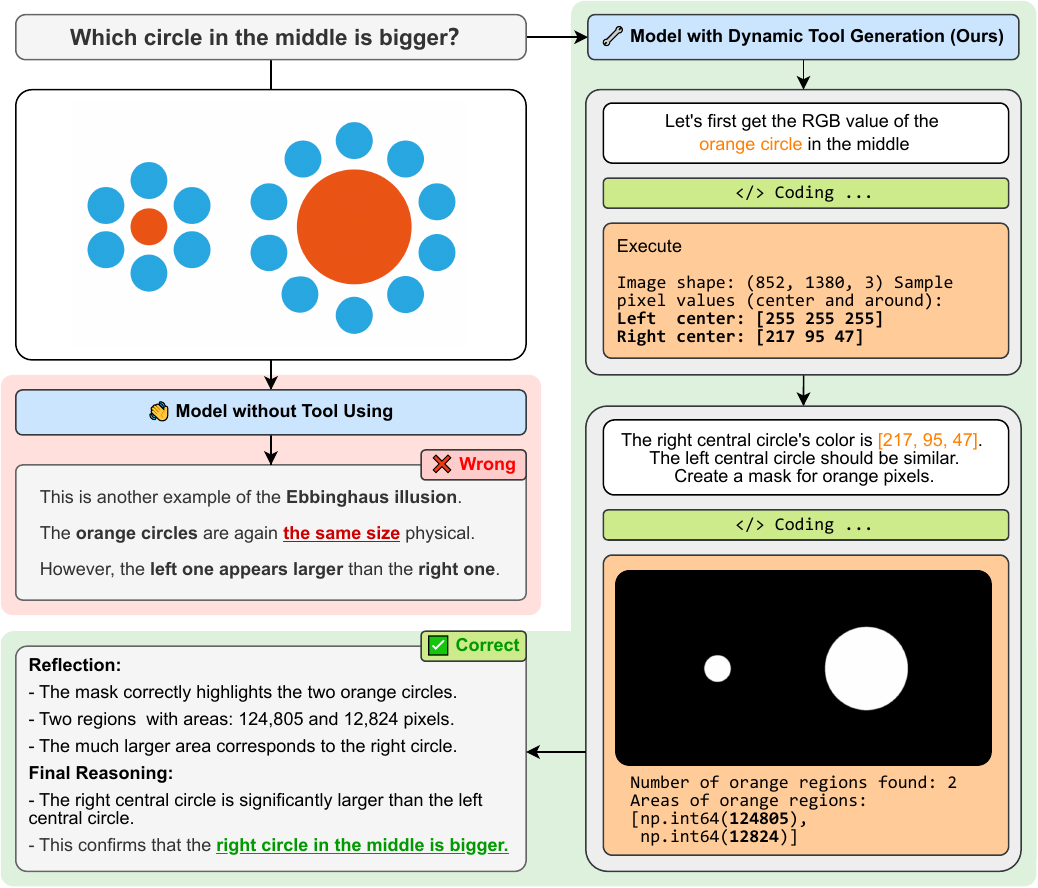}
    \caption{\textbf{Overcoming Illusory Heuristics with Code.} This visual puzzle mimics the well-known Ebbinghaus illusion~\citep{jaeger2015ebbinghaus}, but with a twist: it reverses the typical size context, making the correct answer visually obvious to humans. Yet, a standard MLLM~\cite{gpt4.1} mistakenly recalls the well-documented illusion template to answer ``same size''. In contrast, \model\ behaves agentically, probing pixel values, segmenting objects, and computing the actual sizes via on-the-fly Python code to reach the correct answer. This example highlights how dynamic tooling enables adaptive, grounded, verifiable visual reasoning beyond superficial pattern matching.}
    \label{fig:ebbinghaus}
\end{figure}



\setcounter{footnote}{0}  
\renewcommand{\thefootnote}{\arabic{footnote}}  


\begin{abstract}


LLMs are increasingly deployed as agents, systems capable of planning, reasoning, and dynamically calling external tools. However, in visual reasoning, prior approaches largely remain limited by predefined workflows and static toolsets. In this report, we present \model, an interactive, multi-turn framework that enables MLLMs to autonomously generate, execute, and refine Python-based tools tailored to the task at hand, unlocking flexible and interpretable problem-solving. We develop a taxonomy of the tools created by \model\ and analyze their usage across a diverse set of benchmarks. Quantitatively, \model\ achieves consistent performance gains, boosting GPT-4.1 by +7.8\% on V* and Claude-4.0-Sonnet by +31.1\% on VLMsAreBlind-mini. These results point to a broader shift: dynamic tooling allows models not just to use tools, but to invent them, advancing toward more agentic visual reasoning.

\end{abstract}

\abscontent


\vspace{25pt}
\section{Introduction}
\vspace{5pt}
\label{sec:intro}

The idea of AI agents, systems that can autonomously plan and execute tasks, is rapidly gaining traction in modern AI research. Large language models (LLMs), originally built for text generation, have quickly evolved into capable agents that can formulate plans, interact with environments, and call external tools or functions to solve complex problems with minimal human oversight~\citep{codex-1,cua,minimaxagent,deepswe2025,qiu2025alita,Manus,genspark,hu2025owl,hong2023metagpt,backlund2025vending,lu2025octotools,kimik2}. But beyond simply using tools, the more profound leap lies in an agent’s ability to invent them, such as dynamically generating code snippets tailored to its task or environment. This capacity to create problem-solving tools on the fly is not just powerful, but foundational to intelligence. As Benjamin Franklin remarked, ``Man is a tool-making animal''.

Interestingly, the idea of using external computational modules for complex reasoning is not new, particularly in the vision domain. Early works such as Neural Module Networks~\citep{neuralmodulenetwork} introduced a parser that orchestrated a set of predefined functions, embracing a neuro-symbolic approach to visual reasoning. This line of work inspired a series of influential successors (\cref{tab:comp_visual_prog}). Unlike end-to-end models, these systems explicitly represent each reasoning step and producing transparent and inspectable intermediate outputs, offering a promising path for tackling complex and compositional visual reasoning.

However, prior works typically rely on predefined workflows and static toolsets within single-turn frameworks, limiting the flexibility, creativity, and adaptability that modern LLM agents can achieve through dynamic tooling. With the growing coding and reasoning capabilities of today’s MLLMs, we can now move beyond these constraints in visual reasoning: models can dynamically generate code snippets in a multi-turn setup, building tools on the fly that are tailored to the task at hand.

Recent developments like OpenAI’s ``Thinking with Images''~\citep{thinkwithimage} highlight this potential, but they offer limited visibility into how this process actually works. In this report, we present and analyze how advanced MLLMs with strong coding abilities, in our case, GPT-4.1~\citep{gpt4.1} and Claude-4.0-Sonnet~\citep{claude4}, can dynamically create and leverage Python-based visual tools. We introduce \model, an interactive framework in which the model autonomously generates, executes, and iteratively refines Python code in response to multimodal user queries. To support this dynamic tooling loop, we build on Python’s rich ecosystem of mature libraries and carefully engineer both the system prompts and the runtime environment to enable seamless, multi-turn interaction between the MLLM and Python interpreter.

We then analyze the tools generated by \model\ in depth. To do so, we construct a taxonomy that classifies the tools into four broad categories: basic image processing, advanced image processing, visual prompting and sketching, and numerical and statistical analysis, alongside a long tail of creative, task-specific operations (\cref{fig:ebbinghaus}). This framework enables us to examine how different benchmarks and domains elicit distinct patterns of tool usage. For instance, perception-heavy tasks often trigger operations like cropping and contrast enhancement, while math and logic benchmarks rely more on visual sketching and numerical analysis. These findings highlight the power of dynamic tool generation: it equips the model with the flexibility to adapt its strategy to the unique demands of each task and domain.

\vspace{-5pt}
\begin{table}[tbp]
    \centering
    \adjustbox{max width=\textwidth}{
    \begin{tabular}{l|ccc}
    \toprule
    Methods & Dynamic Workflow & Dynamic Tool Generation & Multi-Turn Framework \\
    \midrule
    NMN~\citep{neuralmodulenetwork} & \texttimes & \texttimes & \texttimes \\
    IEP~\citep{IEP} & \texttimes & \texttimes & \texttimes \\
    VisProg~\citep{visprog} & \texttimes & \texttimes & \texttimes \\
    Visual ChatGPT~\citep{visualchat} & \texttimes & \texttimes & \checkmark \\
    ViperGPT~\citep{23viper} & \texttimes & \texttimes & \texttimes \\
    MM-REACT~\citep{mmreact} & \texttimes & \texttimes & \texttimes \\
    HuggingGPT~\citep{hugginggpt} & \texttimes & \texttimes & \texttimes \\
    Image-of-Thought~\citep{iot} & \texttimes & \texttimes & \texttimes \\
    Visual Sketchpad~\citep{visualsketchpad} & \checkmark & \texttimes & \checkmark \\
    VAT~\citep{vat} & \texttimes & \texttimes & \texttimes \\
    \midrule
    \model\  & \checkmark & \checkmark & \checkmark \\
    \bottomrule
    \end{tabular}}
    \vspace{-5pt}
    \caption{\textbf{Comparison} between \model\ and previous tool-using methods for visual reasoning.}
    \vspace{-10pt}
    \label{tab:comp_visual_prog}
\end{table}

Results across major benchmarks reveal that \model\ consistently improves the performance of strong backend models. Notable improvements include a +7.8\% boost on V*~\citep{wu2024vstar} with \model-GPT-4.1, an +8.3\% gain on Visual Puzzles~\citep{song2025visualpuzzles}, and a dramatic leap on VLMsAreBlind-mini~\citep{vlmsareblind}, where \model-Claude-4.0-Sonnet improves from 48.1\% to 79.2\%, marking a remarkable +31.1\% increase. Our results suggest that \model\ acts as an amplifier of the backend model's innate strengths: gaining more at perception tasks when paired with perceptually strong models like GPT-4.1, and at abstract reasoning when paired with Claude-4.0-Sonnet. In short, dynamic tooling does not override model capabilities. It unlocks them.

Ultimately, the agentic \model\ with dynamic tooling not only provides practical performance benefits, it also signals a broader shift in multimodal reasoning. By empowering models to invent new computational tools on the fly, we move closer to versatile, autonomous, and genuinely creative AI systems capable of adapting in real-world visual reasoning scenarios.

\vspace{-3pt}
\section{PyVision}
\vspace{-2pt}
\label{sec:pyvision}

We propose \model, an interactive, multi-turn framework for multimodal reasoning. \model\ empowers an MLLM with the ability to dynamically generate and execute Python code during inference. In each session, the MLLM receives an input, generates Python code in response, and executes it within an isolated Python runtime. The resulting output—textual, visual, or both—is fed back into the MLLM's context, allowing it to iterate and refine its reasoning over multiple turns until it produces a final answer.

Unlike prior approaches that rely on a fixed toolset, such as detection~\citep{groundingdino} or segmentation~\citep{kirillov2023sam} models, \model\ provides only Python as building blocks for tools. This design leverages Python’s rich ecosystem of scientific and vision libraries, for example, OpenCV~\citep{opencv}, Pillow~\citep{pillow}, NumPy~\citep{numpy}, Pandas~\citep{pandas}, 
Scikit-learn~\citep{scilearn}, 
and Scikit-image~\citep{sciimage}. With access to such a versatile ecosystem, the model can generate highly adaptive tools tailored to diverse tasks.

\paragraph{System Prompt Design.}
To guide the MLLM’s reasoning and code generation, \model\ uses a carefully constructed system prompt in addition to user queries. The system prompts encode operational instructions that specify how to access input images, structure code, and return final answers. Key components include:

\vspace{-10pt}
\begin{itemize}
\item Encouraging the MLLM to generate code to solve the task.
\item Input images or video frames are pre-loaded as variables named \texttt{image\_clue\_i}, where \texttt{i} denotes the image index. This allows the model to reference the images without additional loading code. We also provide image resolution that helps operations like cropping.

\item Output from the code is expected via specific functions: \texttt{print()} for textual results and \texttt{plt.show()} for image visualizations.

\item Each generated code block is wrapped in a \texttt{<code>} tag to enable reliable parsing.

\item Final answers are enclosed in a \texttt{<answer>} tag for consistent evaluation.
\end{itemize}
\vspace{-10pt}

\begin{figure}[tbp]
\centering
\includegraphics[width=1\textwidth]{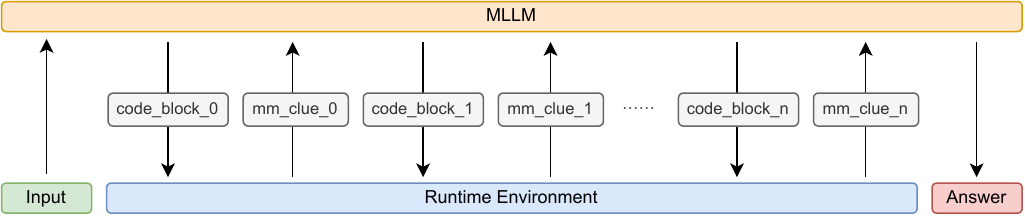}
\caption{\model, an interactive and multi-turn framework capable of dynamic tool generation, designed for multimodal reasoning. In an inference session, \model\ performs \texttt{n+1} interaction turns with the Python interpreter. In the figure, \texttt{code\_block\_i} refers to the generated Python code by the MLLM in the \texttt{i}-th turn, and \texttt{mm\_clue\_i} the executed multi-modal outputs by the Python interpreter. This loop continues until the MLLM outputs a final answer.}
\label{fig:methods}
\end{figure}

With this design, the two MLLMs we experiment with, GPT-4.1~\citep{gpt4.1} and Claude-4.0-Sonnet~\citep{claude4}, can reliably generate parsable and executable code blocks that rarely crash. The full system prompt is included in \cref{sec:prompt_appendix}.

\paragraph{Multi-Turn Interaction between Runtime and the MLLM.}
As illustrated in \cref{fig:methods}, \model\ operates as a multi-turn agentic loop between the MLLM and an isolated Python runtime. In the \texttt{i}-th turn, the MLLM generates a code block \texttt{code\_block\_i}, which is executed to produce multimodal results \texttt{mm\_clue\_i}. These results are appended to the MLLM’s context, enabling it to update its reasoning in the next turn. This loop continues until the MLLM automatically decides to output a final boxed answer.

To support robust and effective multi-turn interaction between the MLLM and the runtime environment of Python, \model\ incorporates several design principles:

\vspace{-10pt}
\begin{itemize}

\item \textbf{Process isolation}: Each code snippet is executed in a subprocess dynamically spawned by the main process, ensuring that crashes or side effects in one execution do not impact the overall inference session.

\item \textbf{Cross-turn persistence}: The runtime environment retains variables and state across turns. This allows the model to reuse or modify intermediate Python code execution results in previous turns, \eg first cropping an image, then applying filters, and finally computing geometric features to complete a task.

\item \textbf{File-system safe I/O}: Communication between the runtime and the MLLM is handled through structured variable passing~\cite{feng2025retool,xue2025simpletir,gou2023tora}, guided by system prompts. This avoids direct dependencies on the host file system.
\end{itemize}
\vspace{-10pt}

Together, these mechanisms enable \model\ to serve as a flexible, secure, and powerful platform for dynamic tool generation in multi-modal reasoning tasks.
\section{Dynamically Generated Tools}
\label{sec:analysis}
\paragraph{Examples in Different Tasks and Domains.}
We start our analysis by presenting examples of \model\ across diverse tasks and domains in \cref{fig:case_medical_case,fig:case_symbolic,fig:vis_science_problem,fig:fig_video,fig:find_difference}.
These examples illustrate how \model\ autonomously creates task-specific and domain-specific tools tailored to each unique challenge, emerging voluntarily from \model's multi-turn code generation and execution.

\subsection{Tooling Taxonomy}
\label{sec:tool_taxonomy}

To better understand the types of tools generated by \model, we construct a taxonomy based on the code it produces across various tasks and domains (\cref{sec:exp}). Specifically, we collect the generated code snippets from inference sessions, embed them using \texttt{text-embedding-3-large}~\citep{textembedding} via OpenAI’s API, and cluster the embeddings to identify emergent tool categories. By inspecting and interpreting the resulting clusters, we identify four major classes of tools: (1) \toolbasic{basic image processing}, (2) \tooladvanced{advanced image processing}, (3) \toolprompts{visual prompting and sketching}, (4) \toolnumerical{numerical and statistical analysis}, and (5) long-tailed operations. We detail each below.

\paragraph{\toolbasic{Basic Image Processing.}}
These tools serve as the foundation for visual manipulation and perception. They enable the model to clean, align, and highlight image content in ways that improve downstream reasoning.

\begin{figure}[tbp]
    \centering
    \includegraphics[width=1\textwidth]{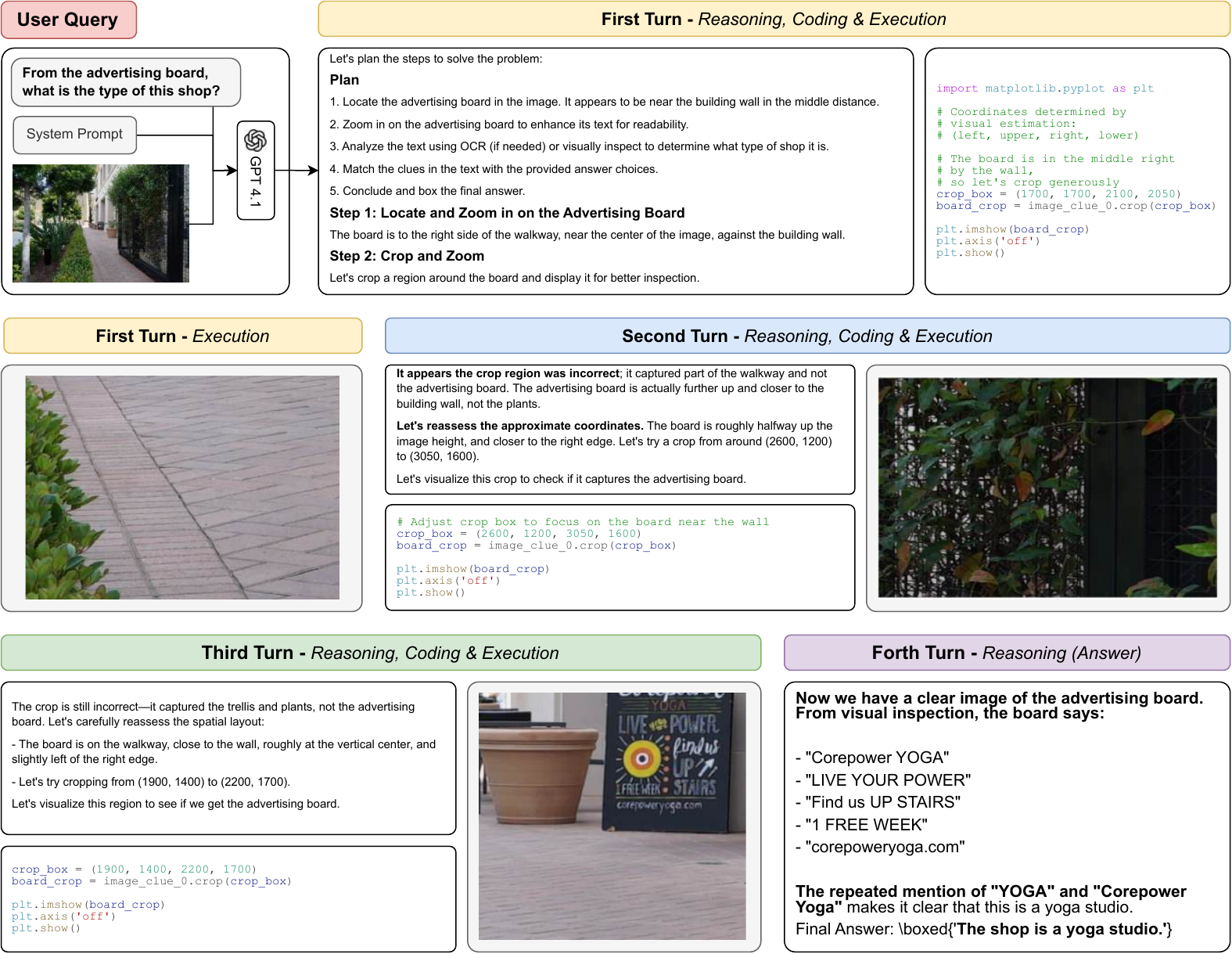}
    \caption{\textbf{Case Study: Visual Search} requires language-based world knowledge, vision-based contextual understanding, and iterative refinement to precisely locate specific visual elements. In this case from V*~\cite{wu2024vstar}, the agent must identify text on a small advertising board in a complex outdoor scene. The target occupies a minimal image area, requiring adaptive cropping. This showcases how dynamic tool generation enables flexible spatial reasoning. \\[5pt]
    The agent generates custom Python codes across three iterative turns, refining coordinate-based \toolbasic{cropping tools} based on visual feedback, and eventually isolating the advertising board required by the user query. The agent then performs \tooladvanced{OCR} and extracts key text such as ``\texttt{YOGA}" and correctly identifies the business as a yoga studio.}
    \label{fig:case_vis_search}
\end{figure}

\begin{figure}[tbp]
    \centering
    \includegraphics[width=1\textwidth]{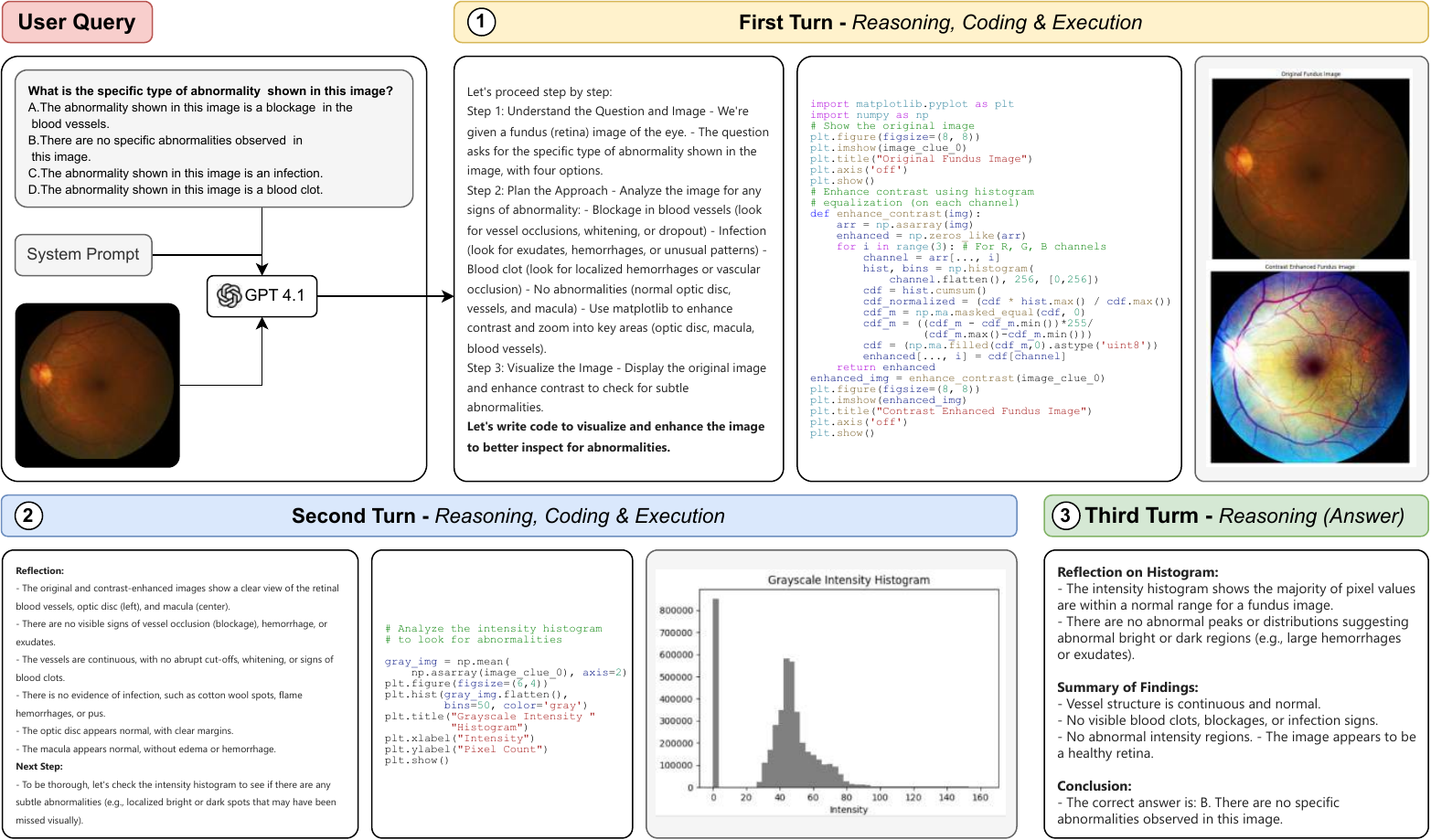}
    \caption{\textbf{Case Study: Medical Imaging Analysis} often requires high visual sensitivity and domain-specific reasoning to solve subtle medical problems. This case from OmniMedVQA~\cite{hu2024omnimedvqa} involves identifying a specific abnormality in a fundus image of the retina.  \model\ is tasked with diagnosing the image without prior domain-specific modules. \\[5pt]
    To begin, \model\ generates a visualization tool that applies histogram equalization to \toolbasic{enhance contrast} in key regions, a standard technique in medical imaging analysis. It then creates a second tool to plot the \toolnumerical{intensity histogram}, confirming the absence of abnormal peaks. Within a few iterations, \model\ constructs a custom diagnostic pipeline from scratch, correctly concluding that no specific abnormalities are present. This highlights \model’s ability to generate interpretable, domain-adapted tools for complex medical tasks.}
        \label{fig:case_medical_case}
\end{figure}

\vspace{-10pt}
\begin{itemize}

\item \toolbasic{Cropping}: For high-resolution or cluttered inputs, \model\ often crops and zooms into regions of interest. By selecting coordinates through reasoning, it effectively performs soft object detection, focusing attention where it matters most. (\cref{fig:case_vis_search})

\item \toolbasic{Rotation}: Misaligned images (\eg rotated maps, skewed documents) can confuse even strong models. \model\ rotates inputs to canonical orientations, making text, spatial layouts, or directional cues easier to interpret.

\item \toolbasic{Enhancement}: In visually subtle domains like medical imaging, \model\ applies contrast adjustments and other enhancements to make latent structures more salient. (\cref{fig:case_medical_case})

\end{itemize}
\vspace{-10pt}

\paragraph{\tooladvanced{Advanced Image Processing.}} These tools reflect \model’s ability to perform mid to high-level vision tasks, but designed and executed dynamically, on demand.

\begin{figure}[tbp]
    \centering
    \includegraphics[width=1\textwidth]{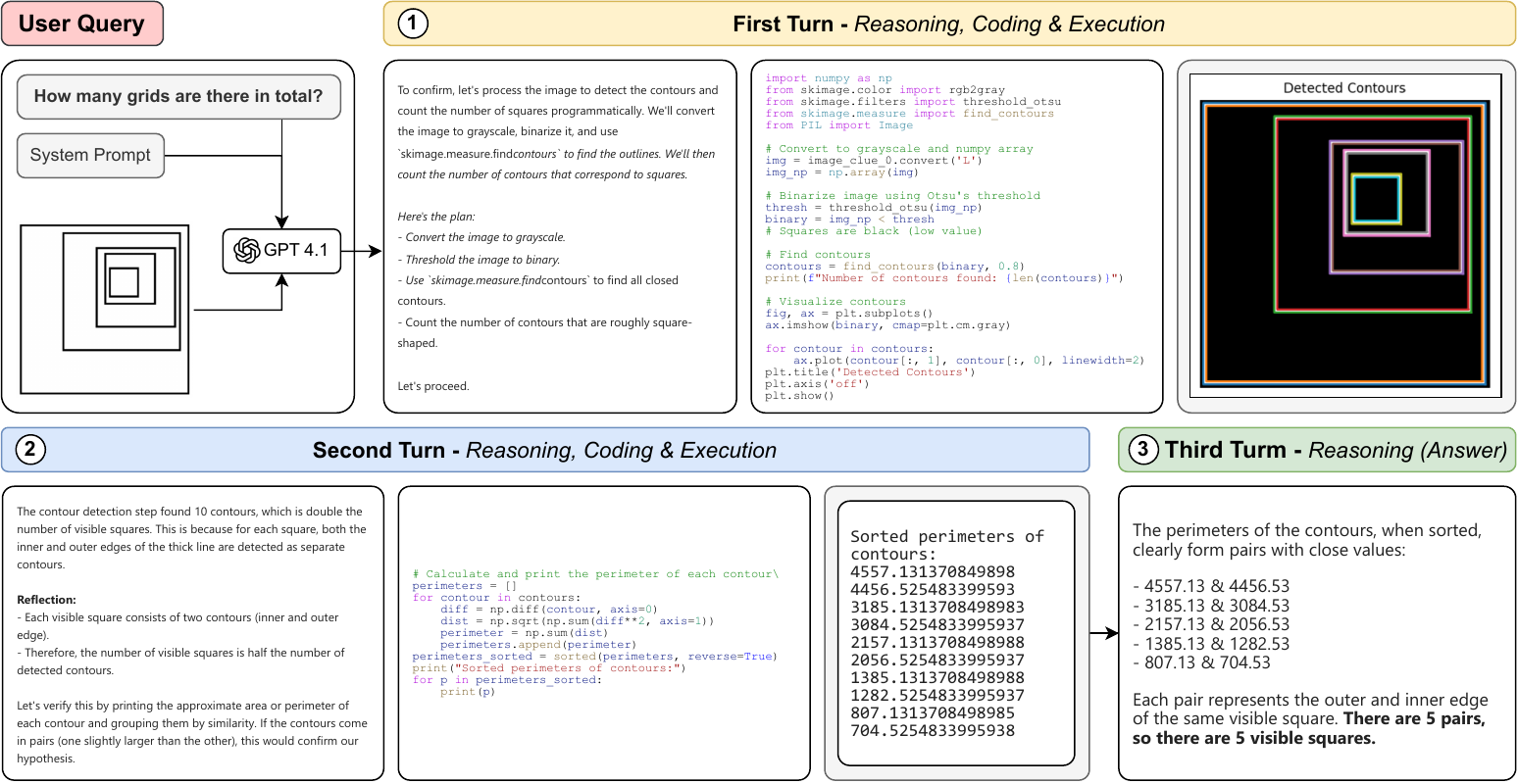}
    \caption{
    \textbf{Case Study: Symbolic Visual Puzzles} require precise spatial reasoning and the recognition of overlapping or closely positioned geometric primitives, a task that is simple for humans, yet surprisingly difficult for powerful MLLMs when asked directly~\cite{vlmsareblind}. In contrast, the regular structure of such puzzles makes them well-suited to code-based solutions. \\[5pt]
    This is a nested squares counting task from VLMsAreBlind~\cite{vlmsareblind}. First, \model\ applies \tooladvanced{edge detection} using \texttt{\detokenize{skimage.measure.find_contours}} and identifies ten contours. Then, it infers that the number of contours corresponds to five nested squares, as each square contributes an inner and outer edge. To validate this, the model performs \toolnumerical{numerical analysis} and prints the sorted perimeters of the detected contours. In the third stage, it performs a double-check and confidently confirms the correct answer: five nested squares.
    }
    \label{fig:case_symbolic}
\end{figure}

\vspace{-10pt}
\begin{itemize}

\item \tooladvanced{Segmentation}: By isolating specific regions via thresholding or edge detection, \model\ can extract foreground objects from background noise.

\item \tooladvanced{Detection}: \model\ generates bounding boxes or edge detection to localize objects in the scene. This supports follow-up operations like counting or measuring. (\cref{fig:case_symbolic})

\item \tooladvanced{OCR}: Without relying on external APIs, \model\ extract textual content (\eg signage, labels) by itself, enabling hybrid visual-linguistic reasoning. (\cref{fig:case_vis_search})

\end{itemize}
\vspace{-10pt}

\paragraph{\toolprompts{Visual Prompting and Sketching.}}
In some tasks, it is not enough to perceive the image—the model must ``think visually''~\citep{som,hong2024cogagent,bai2024digirl,webagent}. To help itself reason, \model\ annotates the image with auxiliary markings, essentially creating visual notes or sketches.

\begin{figure}[tbp]
    \centering
    \includegraphics[width=1\textwidth]{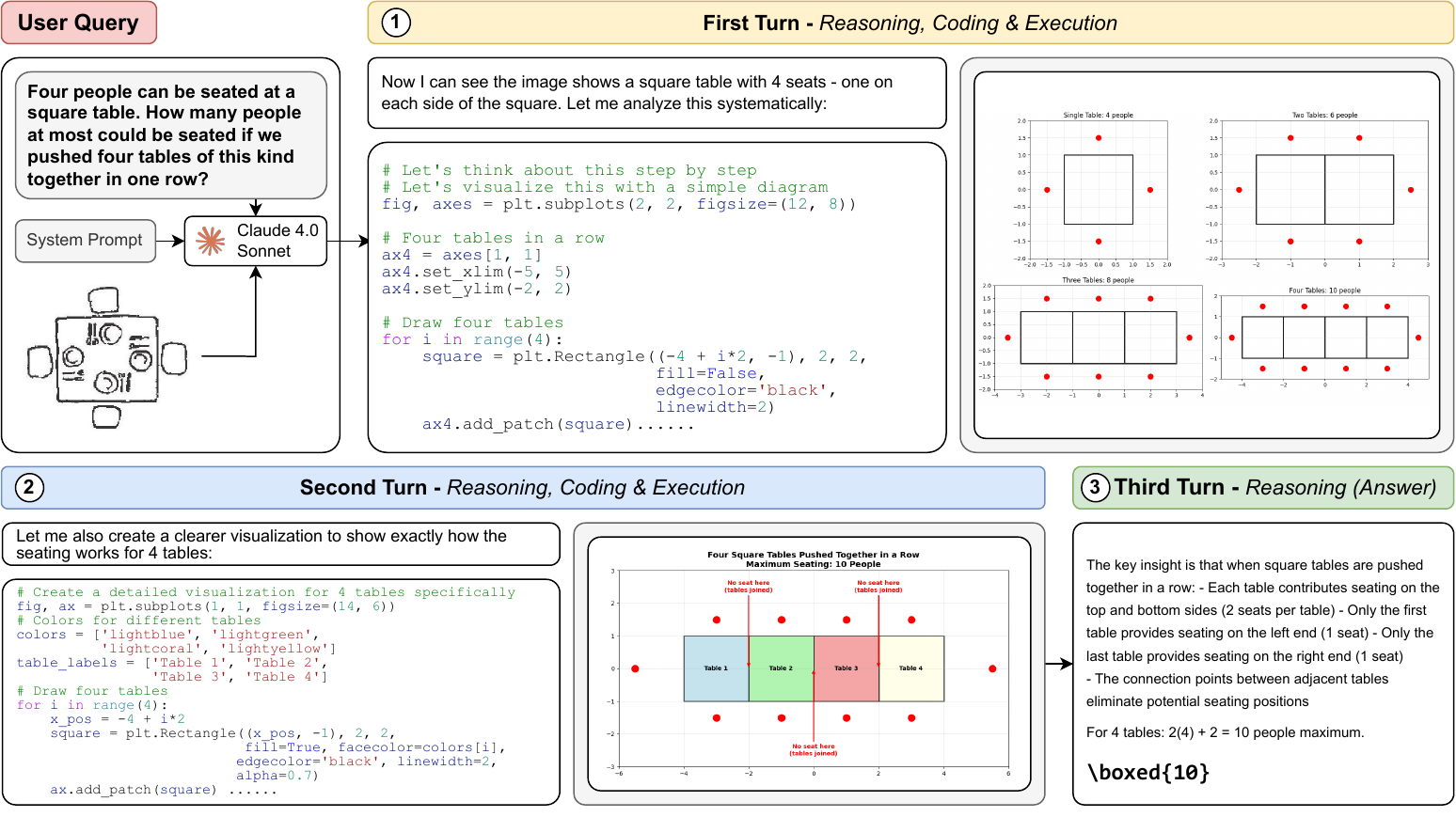}
    \caption{\textbf{Case Study: Visual Sketching} is a valuable strategy for humans to solve scientific problems, including those in mathematics and physics. It can also enhance AI model performance by enabling precise numerical calculations and visual reasoning~\cite{visualsketchpad}.\\[5pt]
    In this example from MathVision~\cite{wang2024mathvision}, \model\ is asked to compute how many people can sit around a row of four tables. The model first sketches the table arrangement, \toolprompts{rendering marks} to represent four connected tables and people sitting around them. From the sketch, it infers that ten people can be seated. In a second turn, \model\ generates a schematic diagram to explain and validate the seating logic, identifying positions that are unsuitable for seating.}
    \label{fig:vis_science_problem}
\end{figure}

\vspace{-10pt}
\begin{itemize} 

\item \toolprompts{Rendering Marks}: In object counting or enumeration task, \model\ often marks items with dots or symbols. This external memory acts as a tallying aid, helping it keep track of what’s been counted. (\cref{fig:vis_science_problem})

\item \toolprompts{Rendering Lines}: In geometric or spatial tasks (\eg mazes), \model\ draws auxiliary lines to assist reasoning, such as showing the moving directions in a maze.

\end{itemize}
\vspace{-10pt}

\paragraph{\toolnumerical{Numerical and Statistical Analysis.}}
To go beyond perception and into interpretation, \model\ invokes tools for quantitative reasoning over visual inputs.

\vspace{-10pt}
\begin{itemize}

\item \toolnumerical{Image Histogram}: By plotting pixel intensity distributions, \model\ can analyze lighting, contrast, and more, critical for domains where histogram carry meaning. (\cref{fig:case_medical_case})

\item \toolnumerical{Numerical Analysis}: When solving visual math problems or compare quantities, \model\ writes scripts to compute areas, lengths, or other metrics for symbolic reasoning. (\cref{fig:case_symbolic})
\end{itemize}
\vspace{-10pt}

\begin{figure}[tbp]
    \centering
    \includegraphics[width=1\textwidth]{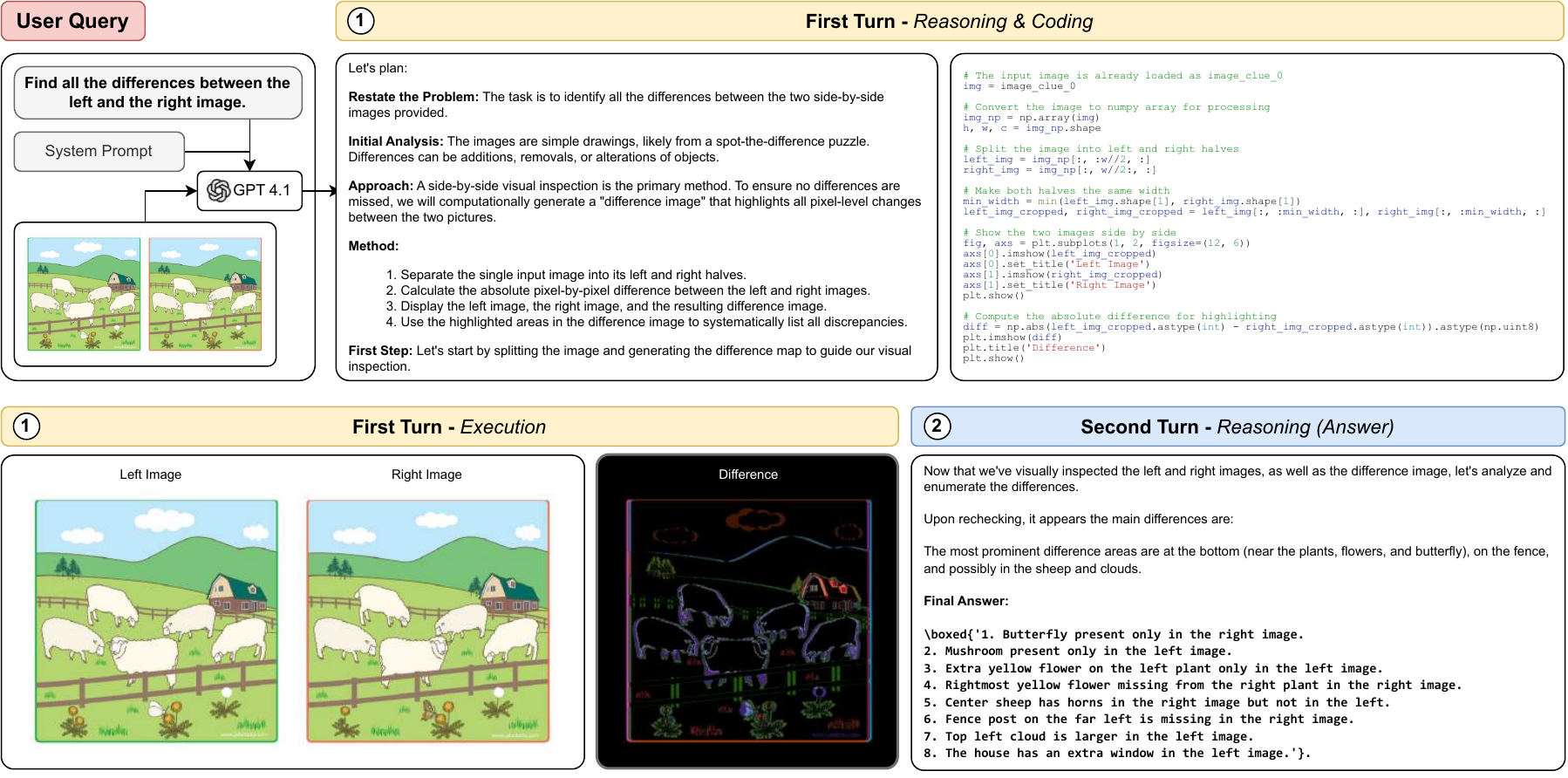}
    \caption[Short caption for list of figures]{\textbf{Case Study: Spot-the-Difference} showcases structured visual comparison. Given a side-by-side image pair,  \model\ is asked to identify all visual discrepancies.\\[5pt]
     \model\ first plans a multi-step strategy: it splits the image into left and right halves, computes the absolute pixel-level difference, and generates a difference map to highlight changes. It then displays both original images alongside the computed difference visualization to aid analysis. Based on this, \model\ proceeds to enumerate the identified differences. Although the final answers are \textit{not completely correct}, the model's initiative to employ pixel-level differencing and organize a reasoning pipeline is notable. This example illustrates both the creative potential of agentic visual reasoning and the ongoing challenge of mitigating hallucinations.%
    \footnotemark}
    \label{fig:find_difference}
\end{figure}

\paragraph{Long-Tail Operations.} \model\ also invents novel tools not easily classified. These one-off operations showcase its ability to reason creatively under novel constraints. For example, in Fig.~\ref{fig:find_difference}, to solve a ``spot the difference'' task, \model\ directly subtracts pixel values between two images and visualizes the result. This kind of zero-shot problem decomposition and tool synthesis reflects both the power and flexibility of dynamic tooling for visual reasoning.

\begin{figure}[tbp]
    \centering
    \includegraphics[width=\textwidth]{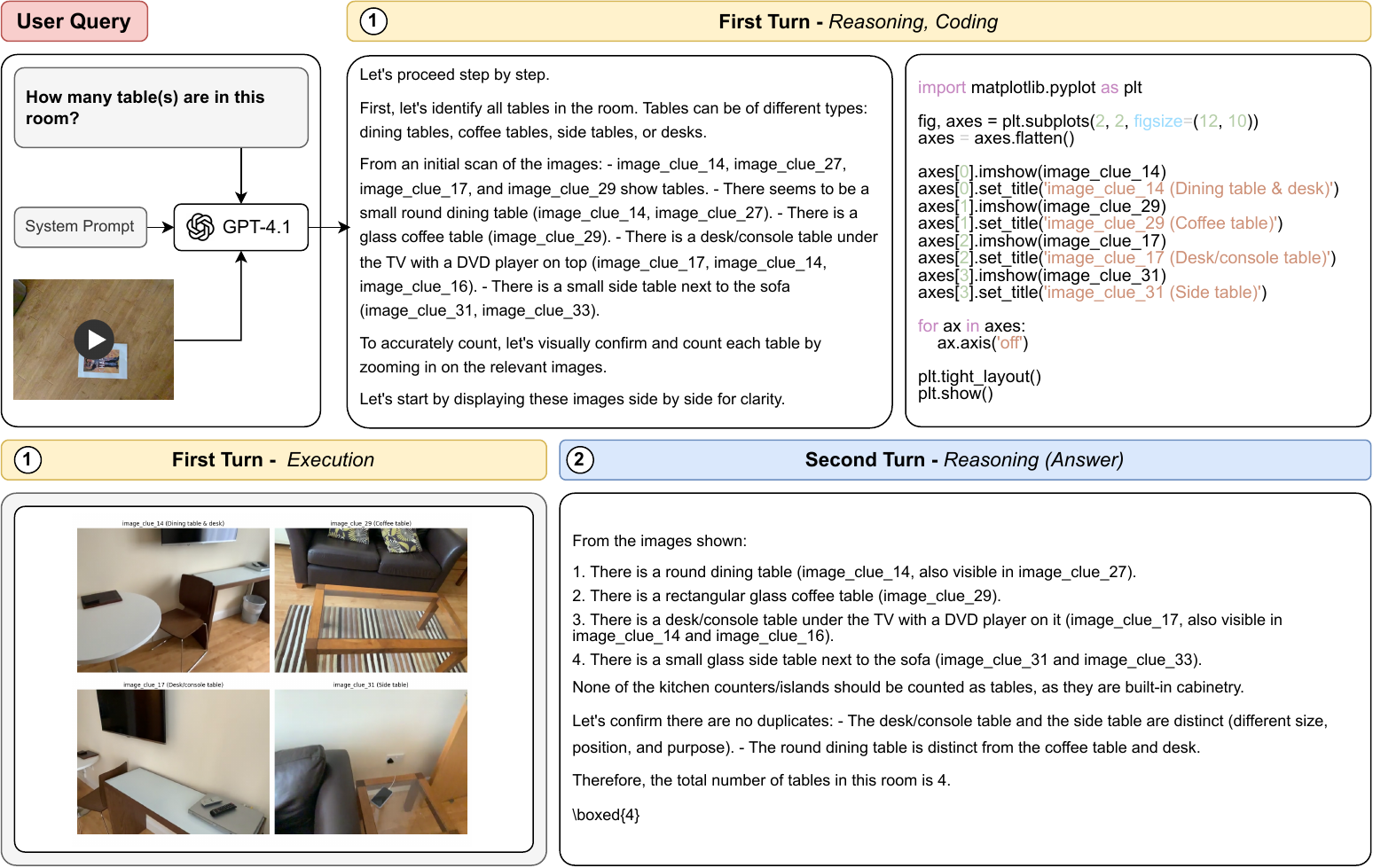}
    \caption{\textbf{Case Study: Video Understanding} benefits from an agentic pipeline that integrates reasoning across frames and targeted frame selection. In this example from VSI-Bench~\cite{yang2024think}, a 3-minute egocentric video of an indoor scene is presented, and \model\ is tasked with identifying the number of tables in the room.\\[5pt]
    \model\ begins by analyzing the video to \tooladvanced{detect} candidate frames containing tables. It then selects and displays key frames, each showing a different table, including dining table, desk, coffee table, and side table, to support its reasoning. By synthesizing visual evidence and textual inference across multiple views, \model\ concludes there are four distinct tables in the room.
    }
    \label{fig:fig_video}
\end{figure}

\paragraph{Video Reasoning with Agentic Tooling.}
Video understanding poses unique challenges compared to static image tasks. \model\ demonstrates strong potential in this setting by treating video not as a monolithic input but as a sequence of decision points. In Fig.~\ref{fig:fig_video}, rather than exhaustively analyzing all frames, \model\ dynamically selects and processes only those frames containing distinct types of tables. It then extracts visual evidence and support reasoning. This agentic, multi-step workflow enables \model\ to operate more like a human analyst: skimming, sampling, and refining its understanding based on intermediate results.

\footnotetext{Data source: \url{https://www.jabobaby.com/blog/posts/photo-hunt}}

\begin{figure}[tbp]
    \centering
    \includegraphics[width=\linewidth]{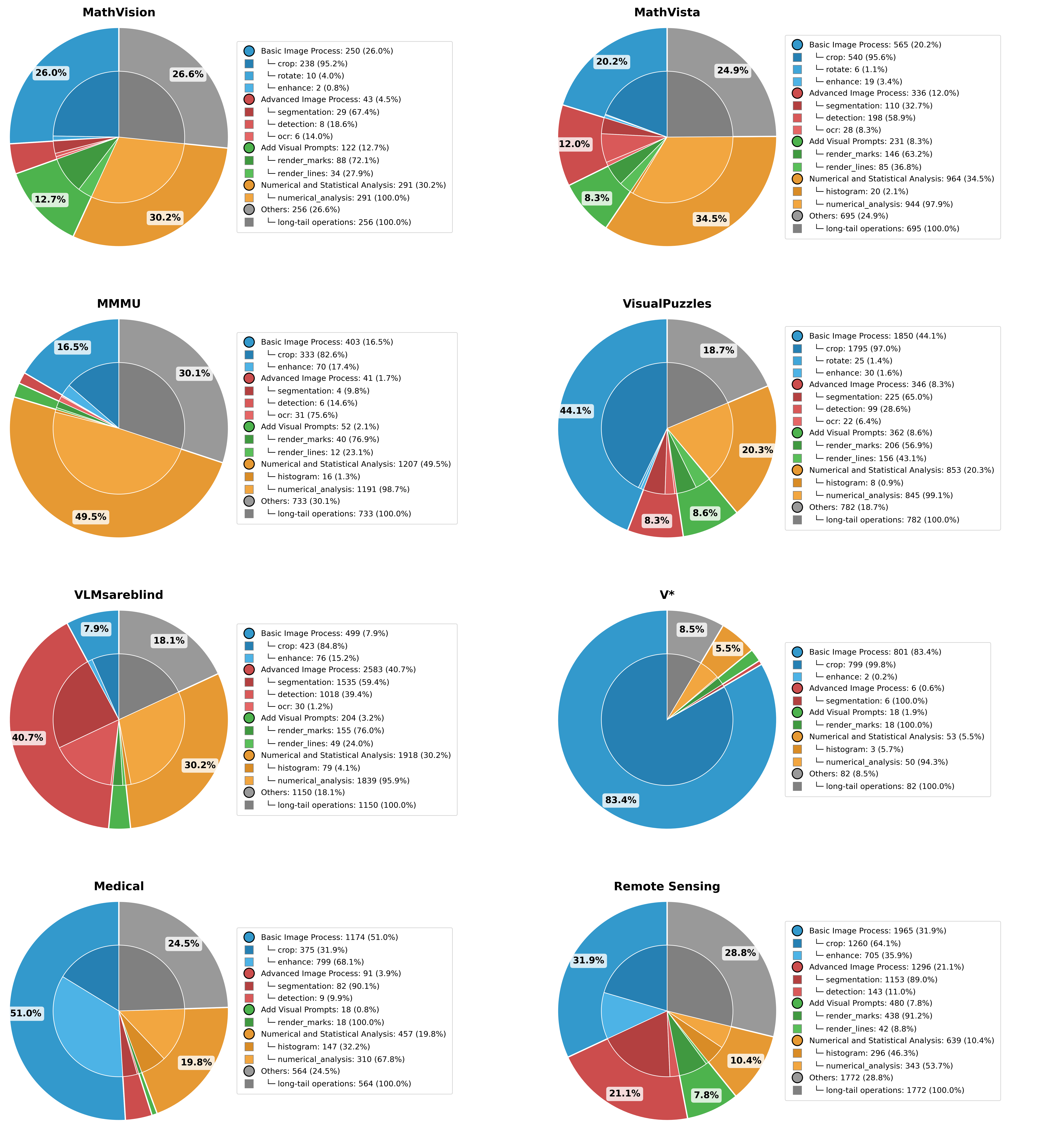}
    \caption{\textbf{Taxonomy Distribution Across Benchmarks and Domains.} Tool usage varies significantly across different \textit{tasks} and \textit{domains}.\\[5pt]
    For math- and logic-related benchmarks, \eg MathVision~\citep{wang2024mathvision}, MathVista~\citep{lu2023mathvista}, MMMU~\cite{yue2024mmmu}, VisualPuzzles~\citep{song2025visualpuzzles}, \toolnumerical{numerical and statistical} tools constitute a major portion of the usage and \toolprompts{visual prompts} are used relatively more often. In the symbolic vision task VLMsAreBlind~\citep{vlmsareblind}, \tooladvanced{advanced image processing} tools dominate. For visual search in V*~\citep{wu2024vstar}, \model\ primarily relies on \toolbasic{cropping} to facilitate detailed visual querying, which takes over 83\% of all tools used.\\[5pt]
    Tooling preferences are also domain-sensitive: On medical images~\citep{hu2024omnimedvqa}, \toolbasic{contrast-enhancement tools} are frequently invoked. In remote sensing~\citep{kuckreja2024geochat}, \tooladvanced{segmentation tools} are more common.\\[5pt]
    These observations highlight the importance of flexible and dynamic tooling to support the diverse demands of real-world vision tasks.}
    \label{fig:tools_taxonomy}
\end{figure}

\subsection{Analyzing Tooling Patterns Across Tasks and Domains}

\paragraph{Benchmarks.} 
To evaluate the effectiveness of \model\ on versatile benchmarks and domains, we select six benchmarks. The details are listed as follows:

\vspace{-10pt}
\begin{itemize}
\item \textbf{Multi-Modal Math}: MathVista~\citep{lu2023mathvista} and MathVision~\citep{wang2024mathvision} challenge models with math problems that combine visual perception and numerical reasoning.

\item \textbf{Domain and Logic Reasoning}: MMMU~\citep{yue2024mmmu} tests subject-specific reasoning across disciplines using multi-modal input, often requiring college-level knowledge. VisualPuzzles~\citep{song2025visualpuzzles} focuses on logic, with tasks covering algorithmic, analogical, deductive, inductive, and spatial reasoning, minimizing domain dependency while maximizing abstraction.

\item \textbf{Symbolic Vision}: VLMs Are Blind~\citep{vlmsareblind} consists of designed symbolic visual puzzles, probing the limits of parsing and reasoning over abstract, structured visual primitives.

\item \textbf{Fine-Grained Visual Search}: V*~\citep{wu2024vstar} features 191 high-resolution samples that require pinpointing subtle visual details based on nuanced queries, making it a strong testbed for attention and spatial reasoning.
\end{itemize}
\vspace{-10pt}

We also evaluate two special domains, Medical Imaging VQA~\citep{hu2024omnimedvqa} and Remote Sensing VQA~\citep{kuckreja2024geochat} to probe the tooling patterns in different domains.

\paragraph{Distribution of Tools.} To understand how \model\ adapts its tooling to different problems, we analyze the distribution of tool categories across benchmarks and domains in \cref{fig:tools_taxonomy}.

The results reveal strong task- and domain-specific preferences. In math and logic-heavy benchmarks like MathVista~\citep{lu2023mathvista}, MathVision~\citep{wang2024mathvision}, MMMU~\cite{yue2024mmmu}, and VisualPuzzles~\citep{song2025visualpuzzles}, \model\ frequently generates \toolnumerical{numerical and statistical} tools to support symbolic and quantitative reasoning. These are often accompanied by \toolprompts{visual prompting and sketching} that help ground abstract logic in visual cues. In symbolic visual tasks such as VLMsAreBlind~\citep{vlmsareblind}, \tooladvanced{advanced image processing} tools are predominant, reflecting the need for structure extraction and visual parsing. For fine-grained visual search tasks like V*~\citep{wu2024vstar}, \toolbasic{cropping} overwhelmingly dominates, accounting for over 83\% of all tools, as the model focuses attention on localized regions.

Domain also plays a significant role: on medical images~\citep{hu2024omnimedvqa}, \toolbasic{contrast enhancement} is commonly used to reveal subtle visual patterns, while in remote sensing~\citep{kuckreja2024geochat}, \tooladvanced{segmentation} tools help delineate objects in large-scale scenes.

These results underscore the importance of dynamic tool generation, allowing the model to flexibly tailor its strategy to the task at hand.




\section{Results on Versatile Benchmarks}
\label{sec:exp}

\paragraph{Baselines.} 
To evaluate \model's effectiveness on diverse multi-modal scenarios, we test it on versatile benchmarks with MLLMs including GPT-4.1~\citep{gpt4.1} and Claude-4.0-Sonnet~\citep{claude4} as the backend.
We use plain chain-of-thought prompting~\citep{wei2022cot,kojima2022lzero-cot} as our baseline. The inference parameter settings and the prompt details are in \cref{sec:prompt_appendix}.


\begin{table}[h]
    \centering
    \adjustbox{max width=\textwidth}{
    \begin{tabular}{l|cccccc}
    \toprule
    & MathVista & MathVision-mini & MMMU & VisualPuzzles & VLMsAreBlind-mini & V* \\
    \midrule
    GPT-4o               & 61.4 & --  & 68.7 & 41.1 & --  & 73.9 \\
    o1                   & 71.8 & --  & 77.6 & 51.8 & -- & 69.7 \\
    o3                   & 86.8 & --  & 82.9 & 54.0 & --  & 95.7 \\
    \midrule
    GPT-4.1              & 69.9\rlap{${}^{*}$} & 46.4 & 71.9\rlap{${}^{*}$} & 44.9 & 67.1 & 68.1 \\
    \model-GPT-4.1       & \scorediff{71.7}{+1.8}
                        & \scorediff{48.7}{+2.3}
                        & \scorediff{74.3}{+2.4}
                        & \scorediff{47.4}{+2.5}
                        & \scorediff{69.7}{+2.6}
                        & \scorediff{75.9}{+\textbf{7.8}} \\
    \midrule
    Claude-4.0-Sonnet    & 71.4 & 48.0 & 74.4 & 42.7 & 48.1 & 56.5 \\
    \model-Claude        & \scorediff{76.2}{+4.8}
                        & \scorediff{51.3}{+3.3}
                        & \scorediff{74.6}{+0.2}
                        & \scorediff{51.0}{+\textbf{8.3}}
                        & \scorediff{79.2}{+\textbf{31.1}}
                        & \scorediff{56.8}{+0.3} \\
    \bottomrule
    \end{tabular}}
    \caption{\textbf{Performance on six benchmarks}. Improvements over each base model appear beneath the scores. We highlight a +\textcolor{green!70!black}{\textbf{7.8}}\% gain on V* by \model-GPT-4.1, +\textcolor{green!70!black}{\textbf{8.3}}\% on VisualPuzzles and +\textcolor{green!70!black}{\textbf{31.1}}\% on VLMsAreBlind-mini by \model-Claude. *GPT-4.1 results are self-collected with plain chain-of-though prompting (\cref{appx:cot-prompt}) in June 2025.}
    \label{tab:benchmark_result}
\end{table}

\paragraph{Results.}
\cref{tab:benchmark_result} highlights how adding \model's dynamic tooling consistently boosts two strong back-end models across a diverse benchmark suite. For GPT-4.1, \model\ yields uniform gains on every dataset, from modest improvements on math-centric tasks: +1.8\% on MathVista and +2.4\% on MMMU, to a sizeable +7.8\% on the fine-grained visual-search benchmark V*. Claude-4.0-Sonnet shows a sharper pattern: while math and general-reasoning tasks improve by roughly +3\% to +5\%, symbolic-vision performance on VLMsAreBlind-mini soars by +31.1\%. In short, dynamic tool generation delivers broad, task-dependent gains, which also depends on the backend model's capability, discussed next.

\paragraph{\model\ Amplifies What the Backend MLLM Does Best, Reasoning or Perception.}
\label{sec:qwen}

To better understand the relationship between \model's performance gains and the inherent strengths of backend models, we focus on two representative benchmarks: \textit{MathVision-mini}~\cite{wang2024mathvision}, which emphasizes abstract reasoning, and \textit{V*}~\cite{wu2024vstar}, which highlights perception ability. Claude-4.0-Sonnet, stronger in abstract reasoning as shown by its higher MathVision-mini performance (48.0\% vs. 46.4\% for GPT-4.1), experiences a larger boost from PyVision (+3.3\%) compared to GPT-4.1’s more modest gain (+2.3\%). Conversely, GPT-4.1, superior in perceptual tasks like V* (68.1\% vs. Claude-4.0-Sonnet's 56.5\%), achieves a significantly greater improvement with \model\ (+7.8\% vs. only +0.3\%). This complementary pattern suggests that the effectiveness of dynamic tooling provided by \model\ depends critically on the backend model's foundational reasoning and perception strengths.

Further supporting this hypothesis, experiments with Qwen2.5-VL-72B~\citep{bai2025qwen25vl} yield similar findings: weaker abstract reasoning capabilities (18.4\% on MathVision-mini) lead to limited improvement (+1.7\%), while stronger perceptual performance (67.0\% on V*) translates into substantial gains (+10.0\%). These insights underline that \model\ amplifies existing backend model strengths, making the interplay of reasoning and perception crucial for unlocking the full potential of dynamic multimodal tooling.

\begin{figure}[tbp]
    \centering
    \vspace{-10pt}
    \includegraphics[width=0.97\linewidth]{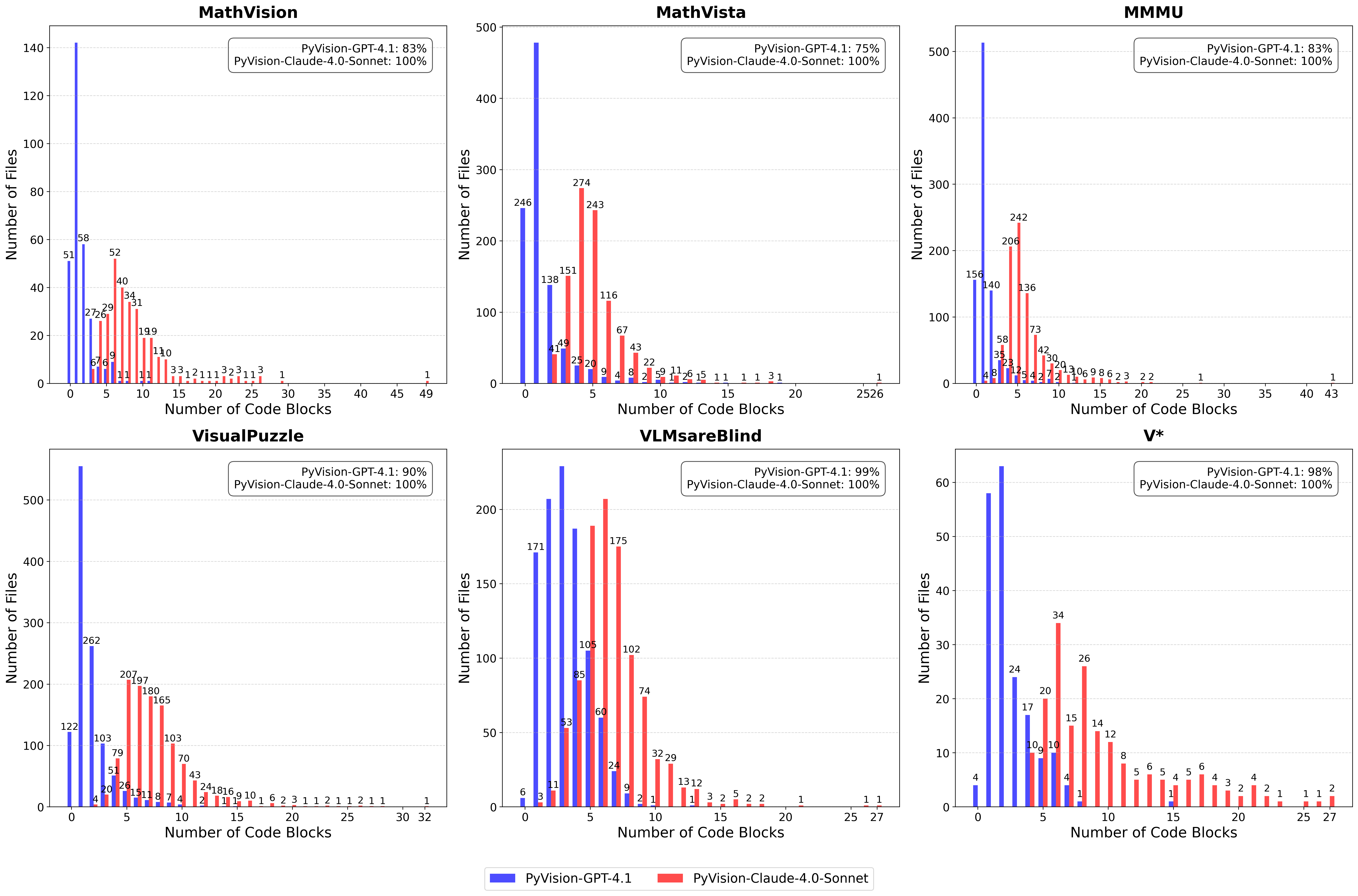}
    \caption{\textbf{Multi-Turn Interaction Patterns Across Tasks and Backend Models}. The histograms show the distribution of the number of generated code blocks per query across six benchmarks. \model-GPT-4.1 (\textcolor{blue}{blue}) and \model-Claude-4.0-Sonnet (\textcolor{red}{red}) exhibit distinct interaction patterns, with Claude consistently generating code more frequently and with more turns. The legend in each subplot indicates the percentage of samples that involved at least one code block.}
    \label{fig:code_dist}
    \vspace{-10pt}
\end{figure}



\paragraph{How Often and How Much MLLMs Generate Code?}
\cref{fig:code_dist} shows the distribution of the number of code blocks generated per user query across six benchmarks, comparing \model\ backed by GPT-4.1 and Claude-4.0-Sonnet. Each subplot visualizes how frequently the model uses code during multi-turn inference, with the legend indicating the percentage of query sessions that include any code generation. We observe that Claude-4.0-Sonnet consistently generates more code than GPT-4.1 across all domains, often with longer toolchains per query and reaching 100\% code coverage. Conversely, GPT-4.1 tends to use fewer code blocks. These trends suggest a difference in agentic behavior, reflecting underlying differences in how each MLLM parses complexity and utilizes code to support reasoning.
\vspace{5pt}
\section{Related Work}
\label{sec:related}

\paragraph{Multi-Modal Tool Using.}
To solve the compositional Visual Question Answering~(VQA) task in a more transparent and interpretable fashion, early work NMN~\citep{neuralmodulenetwork} use a heuristic method while IEP~\citep{IEP} train an LSTM network as the program generator. In the era of LLMs, a pretrained LLM, \eg GPT-4, is used to generate programs.

Visual ChatGPT~\citep{visualchat}, MM-REACT~\citep{mmreact}, HuggingGPT~\citep{hugginggpt}, Image-of-Thought~\citep{iot}, and VAT~\citep{vat} design workflows to process VQA inputs and produce final answers. In VisProg~\citep{visprog} and ViperGPT~\citep{23viper}, researchers predefine a static toolset for specific vision tasks and prompt the LLMs or MLLMs to generate programs that invoke these tools to support reasoning. As LLMs' coding abilities improve, Visual Sketchpad~\citep{visualsketchpad} predefines a toolset and prompts the LLM to program and execute code on the fly, offering more flexibility. These prior works rely on a static toolset containing various visual parsers~\citep{visualparser}, \eg detection models (GroundingDINO~\citep{groundingdino}) and segmentation models (SAM~\citep{kirillov2023sam}), which limits generality across vision tasks and makes the external models a bottleneck. In contrast, \model\ uses Python as the sole primitive tool. With the advanced coding and multimodal understanding abilities of today’s MLLMs, \eg Claude-4.0~\citep{claude4} and GPT-4.1~\citep{gpt4.1}, they can write Python code to construct and execute complex tools on the fly, enabling more general and flexible reasoning.

\paragraph{Thinking with Images.}
In o3's~\citep{thinkwithimage} blog, thinking with images is presented as an attractive feature. CoGCoM~\citep{cogcom} synthesizes program-integrated data and teaches the MLLM to use predefined tools during inference. DeepEyes~\citep{zheng2025deepeyes}, Pixel Reasoner~\citep{su2025pixelreasoner}, OpenThinkIMG~\citep{su2025thinking,su2025openthinkimg}, and Chain-of-Focus~\citep{zhang2025chainoffocus} incentivize MLLMs to develop the ability to ``think with images using predefined tools'' through reinforcement learning. In \model, we support thinking with images by using Python as the tool creation interface, enabling the MLLM to self-generate more complex and adaptive tools based on varying scenarios.

\clearpage
\vspace{-5pt}
\section{Conclusion}
\vspace{-8pt}
\label{sec:conclusion}

We propose \model, an agentic framework enabling MLLMs to generate and execute Python code on the fly. Different from previous visual programming works~\citep{23viper,visprog,visualsketchpad}, \model\ needs no visual parsers and predefined static toolset, it generates tools dynamically from the specific query and visual input. We evaluate its effectiveness and flexibility on various benchmarks and visual reasoning scenarios, \eg medical, multi-modal math problems, remote sensing and visual puzzles. It shows significant performance improvement on versatile benchmarks.

\paragraph{Acknowledgement} 
We thank Yuxiang Lai and Jike Zhong for providing test samples in the initial stage of this project and Yunfei Xie for his feedback on the manuscript.
\bibliography{arxiv} 
\newpage
\appendix
\section*{Appendix Contents}

\noindent \hyperlink{A}{A. Additional Evaluation Details\dotfill 21}

\hyperlink{A1}{A.1. System Prompt Details \dotfill 21}

\hyperlink{A2}{A.2. Evaluation Parameters Details \dotfill 22}

\noindent \hyperlink{B}{B. Examples of Generated Tools \dotfill 22}

\hyperlink{B1}{B.1. Code Snippet of \crop\ Tool \dotfill 22}

\hyperlink{B2}{B.2. Code Snippet of \rotatetool\ Tool \dotfill 22}

\hyperlink{B3}{B.3. Code Snippet of \enhance\ Tool \dotfill 23}

\hyperlink{B4}{B.4. Code Snippet of \seg\ Tool \dotfill 23}

\hyperlink{B5}{B.5.  Code Snippet of \detec\ Tool \dotfill 24}

\hyperlink{B6}{B.6. Code Snippet of \ocr\ Tool \dotfill 24}

\hyperlink{B7}{B.7. Code Snippet of \rendermarks\ Tool \dotfill 25}

\hyperlink{B8}{B.8. Code Snippet of \renderlines\ Tool \dotfill 25}

\hyperlink{B9}{B.9. Code Snippet of \hist\ Tool \dotfill 26}

\hyperlink{B10}{B.10.  Code Snippet of \numericalana\ Tool \dotfill 26}

\newpage

\section{Additional Evaluation Details}
\label{sec:prompt_appendix}
\hypertarget{A2}{}\subsection{System Prompt Details}
\begin{figure}[H]
    \centering
\input{tables/system_prompt}
\end{figure}

\hypertarget{A1}{}\subsection{Evaluation Details}

\paragraph{Inference Parameters.}
\label{appx:cot-prompt}
In the evaluation stage, we set the temperature to 0.6. Here is the chain-of-thought prompt template used for evaluation.

\begin{figure}[H]
    \centering
\input{tables/cot_prompt}
\end{figure}

\paragraph{Illustration of GPT-4.1's result in Tab.~\ref{tab:benchmark_result}.}
To keep the consistent evaluation setting with Claude-4.0-Sonnet, we evaluated GPT-4.1 on MathVista~\citep{lu2023mathvista} and MMMU~\citep{yue2024mmmu} by ourselves with the above-mentioned CoT prompt.

\paragraph{Illustration of Qwen2.5-VL-72B's result in Sec.~\ref{sec:qwen}.}
When evaluating Qwen2.5-VL-72B on V*~\citep{wu2024vstar} and MathVision-testmini~\citep{wang2024mathvision}, to keep the consistent evaluation setting with other models, \eg GPT-4.1 and Claude-4.0-Sonnet, we wrap the visual question with the above-mentioned CoT prompt template as input. Since Qwen2.5-VL-72B was not specifically optimized for CoT reasoning~\citep{bai2025qwen25vl}, the results with CoT prompt may be damaged.

\section{Examples of Generated Tools}

\hypertarget{B1}{}\subsection{Code Snippet of \crop\ Tool}
\label{sec:atarigame_appendix}

\begin{lstlisting}[style=pythonstyle, caption={\textbf{A code snippet example of generated \crop\ tool.} In this example, \model\ was queried to answer the type of the shop based on the advertising board's information. It generated \crop\ tool to focus on the board area.}]
# The board is towards the center-right.
# Based on the apparent position, let's crop the region centered at (center-right), roughly:
# x: 900-1050, y: 520-700 (after visually inspecting the source image proportions)
cropped = image_clue_0.crop((900, 520, 1050, 700))

plt.imshow(cropped)
plt.title('Zoom on Advertising Board')
plt.axis('off')
plt.show()
\end{lstlisting}

\hypertarget{B1}{}\subsection{Code Snippet of \rotatetool\ Tool}
\label{sec:atarigame_appendix}
\begin{lstlisting}[style=pythonstyle, caption={\textbf{A code snippet example of generated \rotatetool\ tool.} In this example, \model\ was queried to answer what is written on an upside-down image. It generated \rotatetool\ tool to make the image rotate 180 degree.}]
import matplotlib.pyplot as plt
import pytesseract
from PIL import Image

# Step 1: Rotate the image 180 degrees to correct orientation
rotated_image = image_clue_0.rotate(180)

# Display the rotated image to verify orientation
plt.imshow(rotated_image)
plt.axis('off')
plt.show()
\end{lstlisting}
\hypertarget{B2}{}\subsection{Code Snippet of \enhance\ Tool}
\label{sec:ablation_rot}
\begin{lstlisting}[style=pythonstyle, caption={\textbf{A code snippet example of the generated \enhance\ tool.} In this example, \model\ was queried to identify retinal abnormalities in a fundus image. It generated the \enhance\ tool to increase image contrast, making subtle pathological features more visible.}]
#Visualize the Image - Display the original image and enhance contrast to check for subtle abnormalities. 

#Let's write code to visualize and enhance the image to better inspect for abnormalities.

def enhance_contrast(img):
    arr = np.asarray(img)
    enhanced = np.zeros_like(arr)
    for i in range(3): # For R, G, B channels
        channel = arr[..., i]
        hist, bins = np.histogram(
            channel.flatten(), 256, [0,256])
        cdf = hist.cumsum()
        cdf_normalized = (cdf * hist.max() / cdf.max())
        cdf_m = np.ma.masked_equal(cdf, 0)
        cdf_m = ((cdf_m - cdf_m.min())*255/
                 (cdf_m.max()-cdf_m.min()))
        cdf = (np.ma.filled(cdf_m,0).astype('uint8'))
        enhanced[..., i] = cdf[channel]
    return enhanced
\end{lstlisting}
\hypertarget{B4}{}\subsection{Code Snippet of \seg\ Tool}
\label{sec:acc_diff_appendix}
\begin{lstlisting}[style=pythonstyle, caption={\textbf{A code snippet example of generated \seg\ tool. }In this example, \model\ was queried to count the number of buildings in a satelite remote sensing image. The entire python tool chain (merged into a single python code snippet for conciseness) generated by \model achieved a similar functionality as segmentation models by thresholding, clustering and filtering step by step.}]
import matplotlib.pyplot as plt
import numpy as np
from skimage.filters import threshold_otsu
from scipy.ndimage import label, binary_opening, binary_closing

# Convert to grayscale
image = image_clue_0
gray = np.array(image.convert('L'))

# Otsu's thresholding on grayscale
otsu_thresh = threshold_otsu(gray)
binary_otsu = gray > otsu_thresh

# Let's apply morphological operations and count the number of buildings.

# Morphological operations to clean up the mask
cleaned = binary_opening(binary_otsu, structure=np.ones((5,5)))
cleaned = binary_closing(cleaned, structure=np.ones((7,7)))

# Label connected components
labeled, num_features = label(cleaned)

print("Number of detected buildings:", num_features)

# Calculate area of each region
region_sizes = np.bincount(labeled.flatten())
min_building_area = 1000  # Adjust as needed for urban scale

# Ignore the background (label 0)
num_buildings = np.sum(region_sizes[1:] > min_building_area)

print("Number of buildings (filtered by area):", num_buildings)
\end{lstlisting}

\hypertarget{B1}{}\subsection{Code Snippet of \detec\ Tool}
\label{sec:atarigame_appendix}
\begin{lstlisting}[style=pythonstyle, caption={\textbf{A code snippet example of generated \detec\ tool.} In this example, \model\ was queried to recognize a small cup of water in the image. It generated \detec\ tool to help cropping.}]
# The table is roughly at the center, estimate the coordinates to crop
# Image size: 1920x930, table is roughly between (700, 250) and (1200, 700)
left, top, right, bottom = 700, 250, 1200, 700
table_area = image_clue_0.crop((left, top, right, bottom))
\end{lstlisting}
\hypertarget{B6}{}\subsection{Code Snippet of \ocr\ Tool}
\label{sec:atarigame_appendix}
\begin{lstlisting}[style=pythonstyle, caption={\textbf{A code snippet example of generated \ocr\ tool.} In this example, \model\ was queried to answer what is the text on the image. It generated \ocr\ tool , using EasyOCR to recognize the text on the image.}]
import easyocr
import numpy as np

# Convert rotated PIL image to numpy array for easyocr
img_np = np.array(rotated_img)

# Initialize EasyOCR reader
reader = easyocr.Reader(['en'])

# Perform OCR
result = reader.readtext(img_np, detail=0)

# Print the recognized text
print(result)
\end{lstlisting}
\hypertarget{B7}{}\subsection{Code Snippet of \rendermarks\ Tool}

\begin{lstlisting}[style=pythonstyle, caption={\textbf{A code snippet example of generated \rendermarks\ tool.} In this example, \model\ was queried to count the people wearing solid red shirts. It generated a \rendermarks\ tool, labeling the people wearing red shirts.}]
# Based on my visual analysis, I can identify the following people in solid red shirts:
# Let me mark them with circles and numbers

# Person 1: Far left, standing person in bright red shirt
plt.plot(45, 140, 'yo', markersize=15, markeredgecolor='yellow', markeredgewidth=3)
plt.text(45, 120, '1', fontsize=16, color='yellow', fontweight='bold', ha='center')

# Person 2: Left side, sitting person in red top
plt.plot(110, 195, 'yo', markersize=15, markeredgecolor='yellow', markeredgewidth=3)
plt.text(110, 175, '2', fontsize=16, color='yellow', fontweight='bold', ha='center')

# Person 3: Center-right area, person in red shirt sitting
plt.plot(380, 200, 'yo', markersize=15, markeredgecolor='yellow', markeredgewidth=3)
plt.text(380, 180, '3', fontsize=16, color='yellow', fontweight='bold', ha='center')

plt.axis('off')
plt.show()
\end{lstlisting}

\hypertarget{B8}{}\subsection{Code Snippet of \renderlines\ Tool}
\label{sec:atarigame_appendix}
\begin{lstlisting}[style=pythonstyle, caption={\textbf{A code snippet example of generated \renderlines\ tool.} In this example, \model\ was queried to answer what is the position after the next step. It generated \renderlines\ tool to help grounding the position via sketching.}]
import matplotlib.pyplot as plt
import matplotlib.patches as patches

# Load image
img = image_clue_0

# Set up plot
fig, ax = plt.subplots(figsize=(8, 6))
ax.imshow(img)

# Green square location (col, row): (2, 4) in 0-based grid (counting from top left)
# Move left 1 step: (1, 4)
start_col, start_row = 2, 4
end_col, end_row = start_col - 1, start_row

# Convert to image coordinates
# Each cell: width = 640/5 = 128, height = 480/5 = 96
cell_w, cell_h = 128, 96

# Center of starting cell
start_x = start_col * cell_w + cell_w // 2
start_y = start_row * cell_h + cell_h // 2
# Center of ending cell
end_x = end_col * cell_w + cell_w // 2
end_y = end_row * cell_h + cell_h // 2

# Draw arrow from green square to left
ax.arrow(start_x, start_y, end_x - start_x, end_y - start_y, 
         head_width=20, head_length=20, fc='red', ec='red', length_includes_head=True)

# Add annotation at end point
ax.text(end_x, end_y, 'HERE', color='red', fontsize=16, fontweight='bold', ha='center', va='center')

plt.axis('off')
plt.show()
\end{lstlisting}
\hypertarget{B9}{}\subsection{Code Snippet of \hist\ Tool}
\label{sec:atarigame_appendix}
\begin{lstlisting}[style=pythonstyle, caption={\textbf{A code snippet example of generated \hist\ tool.} In this example, \model\ was queried to if there is some specific pattern. It generated \hist\ tool to help detect the pattern.}]
# Plot the histogram of pixel values
img_array = np.array(image_clue_0)
plt.figure(figsize=(6,4))
plt.hist(img_array.ravel(), bins=256, color='gray')
plt.title('Pixel Value Histogram')
plt.xlabel('Pixel Intensity')
plt.ylabel('Frequency')
plt.show()
\end{lstlisting}
\hypertarget{B10}{}\subsection{Code Snippet of \numericalana\ Tool}
\label{sec:atarigame_appendix}
\begin{lstlisting}[style=pythonstyle, caption={\textbf{A code snippet example of generated \numericalana\ tool.} In this example, \model\ was queried to answer the the color of the specific area in the image. It generated \numericalana\ tool to help analysis the color.}]
import numpy as np

# Convert the glass area to a numpy array
glass_np = np.array(glass_area)

# Calculate mean RGB values
mean_rgb = glass_np.mean(axis=(0, 1))
print("Mean RGB values of the glass area:", mean_rgb)
\end{lstlisting}

\end{document}